# Hierarchical RNN for Information Extraction from Lawsuit Documents

Rao Xi, and Ke Zhenxing

*Abstract*—Every lawsuit document contains the information about the party's claim, court's analysis, decision and others, and all of this information are helpful to understand the case better and predict the judge's decision on similar case in the future. However, the extraction of these information from the document is difficult because the language is too complicated and sentences varied at length. We treat this problem as a task of sequence labeling, and this paper presents the first research to extract relevant information from the civil lawsuit document in China with the hierarchical RNN framework.

## I. INTRODUCTION

THE extraction of information from the lawsuit document in China is always a difficult problem. First of all, in these documents, each case has a variety of content in real life so that it is hard to standardize the format of the document. Second, the legal language is complicated. The sentences varied in length, and the use of word and phrase are irregular and thus, it is difficult for rule-based methods to recognize and classify the information correctly and efficiently. In addition, the number of lawsuit document is too large and it is impossible for people to collect the information manually.

However, the extraction of information is important for the legal profession. If we can obtain the information about the party's claim, court's analysis and decision, we can understand the lawsuit more quickly and better. In the future, we can predict the judge's decision based on this information. Therefore, we have to conquer this hard problem and fill this gap.

As the technology of natural language process developed quickly in the recent year, this type of problem seems to be solved gradually. For example, Jagannatha and Yu (2016) [1] wanted to extract the medical event and their attribute from the text in the Electronic Health Record (EHR). They encountered similar problems including "incomplete sentences, phrases and irregular use of language" as well as "abundant abbreviations, rich medical jargons, and their variations" which made hard for traditional method to extract the information from the records since it is not easy to build such a system based on hard-coding rules. Today, Recurrent Neural Networks (RNNs) provides an alternative way to solve this problem.

Using the hierarchical RNN framework, we treat our problem as a task of sequence labeling and have successfully extracted the information from civil lawsuit documents which are recorded in Chinese. To our knowledge, we are the first group to reach this goal.[1]

## II. RELATED WORK

Before the RNNs, researchers had used different methods to extract the information from the lawsuit document.

For example, Hachey and Grover (2005) [2] collected 188 House of Lords judgment in the UK and used the naive Bayes and maximum entropy model to capture cue phrase information. With the similar method, Moens et al. (2007) [3] got 68% accuracy in the classification in the argument part of legal texts.

In addition, McCarty (2007) [4] collected 111 federal civil law-suit document from the Appellate court in America and used the statistical parsing method ("deep semantic interpretation") to extract information from court's decision part.

Now, RNNs provide a newer and better option for the text classification and information extraction. In the field of natural language processing, applications of RNNs (Recurrent Neural Networks) have achieved promising results, such as language modeling (Mikolov et al., 2011) [5] and machine translation (Cho et al., 2014) [6]. Long Short-Term Memory Networks (Hochreiter and Schmidhuber, 1997) [7] are presented to handle long-distance dependencies, which has always been the problem of standard RNNs. Hierarchical LSTM is a special version of hierarchical recurrent neural networks (El Hihi et al., 1996) [8] and recently was used to model sentence-paragraph relationship in natural language task (Li, Jiwei et al., 2015) [9].

## III. DATASET

We introduce a novel corpus consisting of lawsuit documents (judgments) from the Chinese court. The dataset including 2153 civil lawsuit documents are collected from China Judgements Online (http://wenshu.court.gov.cn/). There are a certain number of sentences in each lawsuit document. In total, the dataset contains 190,203 labeled sentences (an average of 88.3 sentences per document) and 5,639,407 Chinese word tokens. Statistics for the Dataset are shown in Table 1.

Each sentence is annotated with one of the five domain specific labels: claim (made by part A and/or part B), court's analysis, court's decision and other. Statistics for labels are shown in Table 2.

Manuscript received December 12, 2017; revised January 8, 2018.
Rao Xi is with the Jurtech, Shanghai, 200021 China (corresponding author to provide phone: 86-15900832132; e-mail: raoxi@jurtech.xyz).
Ke Zhenxing is with the Maurer School of Law, Indiana University Bloomington, Bloomington, IN 47405 USA (kezh@umail.iu.edu).

[1] We build a demo of this work which could be found at: http://demo.ikuquan.com/

TABLE I
STATISTICS FOR THE DATASETS

| Data | No. |
|---|---|
| Documents | 2153 |
| Sentences in document (total) | 190203 |
| Sentences in document (mean) | 88.34 |
| Sentences in document (max) | 474 |
| Sentences in document (75%) | 106 |
| Words in sentences (total) | 5639407 |
| Words in sentences (mean) | 29.66 |
| Words in sentences (max) | 519 |
| Words in sentences (75%) | 39 |

TABLE II
LABEL DESCRIPTIONS

| Label | No. |
|---|---|
| Documents | 2153 |
| Sentences in document (total) | 190203 |
| Sentences in document (mean) | 88.34 |
| Sentences in document (max) | 474 |
| Sentences in document (75%) | 106 |
| Words in sentences (total) | 5639407 |
| Words in sentences (mean) | 29.66 |
| Words in sentences (max) | 519 |
| Words in sentences (75%) | 39 |

### A. Claim

The first part is the claim. In this part, the party (A) demanded or asserted his or her legal right, and ask for the compensation, payment or reinstatement. Some-times, the party (B) world file the counterclaim against part (A), and the court consolidated these two lawsuits into one.

### B. Analysis

In court's analysis, the judge applied the relevant statute or legal principle to the fact of the case and discussed which side's claim was more reasonable.

### C. Decision

Based on these analysis, the judge made the conclusion on which claim would be granted and which party would win the lawsuit. The "claim" or "decision" each may be described in one or more sentences. And the length of the sentences varied.

### D. Other Categories

This part contains the information about the background of the lawsuit, evidence each side submitted, and investigations by the court. Sometimes, this part includes some special procedures. For example, the Employment Dispute Mediation and Arbitration Law requires the mandatory employment arbitration to be con-ducted by arbitration agency before the company or the employees bring the case to the court. If either of participants feels dissatisfied with the result of arbitration, they can then submit the case to the court. So, in the employment law case, the information on the employment arbitration would be included.

Sometimes, when the judge wrote the lawsuit document, he or she would give guidance for some information. For instance, when judge wrote the claim part, he or she would write the words like "原锟斤拷锟斤拷锟斤拷" (Plaintiff request) or "锟斤拷锟斤拷锟叫撅拷" (Request a judgment on) to suggest that relevant information would follow. Similar, when judge wrote the court's decision, he or she usually wrote the words "锟斤拷院锟叫撅拷" (The court ruling) before the items of court's decision. However, we cannot find all of the information if we only use these hints because other sentences may belong to this type, but did not have such information.

In this work, our dataset has been randomly split into training, validation, and test sets consisting of 1936, 64, and 153 documents, respectively. The distribution of labels in different datasets is the same, which means the "Other" category takes 93% of the total labels.

## IV. MODEL

### A. Long Short-Term Memory (LSTM)

LSTMs is a special kind of Recurrent Neural Networks (RNNs). The structure of LSTM allows it to effectively handle long-term dependencies (address the vanishing gradient problem), which makes them widely used in natural language processing tasks. A traditional LSTM could be summarized as:

$$f_t = \sigma_g(W_f x_t + U_f h_{t-1} + b_f) \quad (1)$$

$$i_t = \sigma_g(W_i x_t + U_i h_{t-1} + b_i) \quad (2)$$

$$o_t = \sigma_g(W_o x_t + U_o h_{t-1} + b_o) \quad (3)$$

$$c_t = f_t \circ c_{t-1} + i_t \circ \sigma_c(W_c x_t + U_c h_{t-1} + b_c) \quad (4)$$

$$h_t = o_t \circ \sigma_h(c_t) \quad (5)$$

where $c_t$ is the cell state vector, which is the key to maintain context information from the previous time-steps. $f_t$ is the forget gate vector and $i_t$ is the input gate. These two gates control the remembering/forgetting of previous information and acquiring new information. $o_t$ is the output gate vector, $W$ and $U$ are weight matrices for gates and state, $b$ variables are the bias.

### B. Bi-directional LSTM

Standard (unidirectional) LSTM only preserves information of the past time-steps. A Bidirectional LSTM (Bi-directional LSTM) (Graves et al., 2005) [10] consists of two standard LSTMs, a forward LSTM and a backward LSTM.

This structure allows the network to preserve information from both past and future time-steps, which means it can understand the context better. The output is the concatenation of the corresponding states of the forward and backward LSTM.

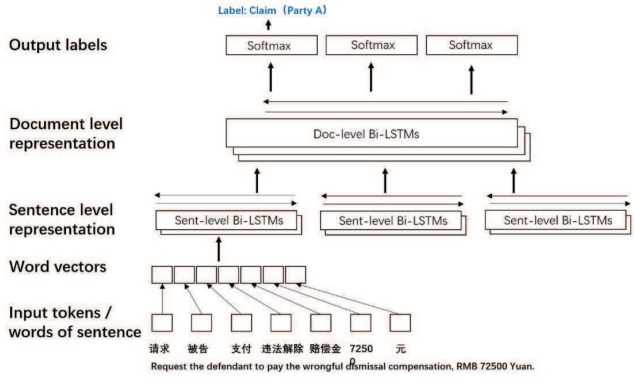

Fig. 1. A hierarchical bidirectional LSTM model.

### C. Hierarchical Bidirectional LSTM

The structure of hierarchical RNN reflects an intuition that the semantics is made up of different levels. The meaning of a document is a joint meaning of the sentences with-in, and it is also the same for a words-sentence relationship.

In this work, to take advantages of the hierarchical structure and the bi-directional structure in NLP task at the same time, we stack document-level Bi-directional LSTMs on top of sentence-level Bi-directional LSTMs, which constitute a Hierarchical Bidirectional LSTM structure as shown in Figure 1.

For any given document (lawsuit document),

$$doc_i = \{sent_k | k = 1...n\} \quad (6)$$

in which $sent_k$ is the $k$th sentence in the document. And for any sentence, each

$$sent_i = \{w_k | k = 1...m\} \quad (7)$$

is consist of a sequence of Chinese words $w$, which tokenized by a commonly used Chinese word segmentation module named "Jieba"[2].

Our model first obtains representation vectors at the sentence level,

$$R_{sent} = BiLSTM(w) \quad (8)$$

by applying a Bi-directional LSTM layer on top of its containing embedded words. The output at the final time-step is used to represent the entire sentence.

We then use a many-to-many multilayer structure (layers of Bi-directional LSTM) to obtains document level representation base on the sentence level representation vectors,

$$R_{doc} = BiLSTM(sent_i) \quad (9)$$

that means the final outputs of sentence level LSTM are fed sequentially as inputs to the first layer of document level BiLSTMs.

Finally, we put hidden state $h_t^{doc}$ of each time-step in the last layer of document level BiLSTMs to a softmax function to produce the final output, label of each sentence:

$$p(sent_k) = Softmax(h_t^{doc}) \quad (10)$$

[2]"Jieba: a Chinese word segmentation module implemented in python," https://github.com/fxsjy/jieba, Accessed: 2017-07-31.

TABLE III
THE PERFORMANCE OF THE METHODS ON THE TEST SET

| Model | F1 | Precision | Recall |
|---|---|---|---|
| BiLSTM with attention (baseline) | 0.0241 | 0.0086 | 0.0814 |
| Hierarchical BiLSTM (1 sent-level and 3 doc-level layers) | 0.3395 | 0.2994 | 0.3921 |
| Hierarchical BiLSTM (2 sent-level and 3 doc-level layers) | 0.356 | 0.3382 | 0.4193 |

## V. EXPERIMENTS

### A. Training

The word embedding matrix is initialized with 256-dimensional pre-trained Chinese Word2Vec vectors (Mikolov et al., 2013) [11]. To prevent overfitting, the pre-trained embeddings are not allowed to be modified during training.

We apply twenty percent of dropout (Srivastava et al., 2014) [12] to both sentence level Bi-directional LSTM layer and document level Bi-directional LSTM layers to avoid over-fitting.

We use categorical cross entropy as the objective function and an RMSProp (Tieleman et al., 2012) [13] optimizer is applied to optimize the network cost.

Batch size is set to 16 (16 documents per batch) during training. We also employ early stopping with a patience of 3 to prevent poor generalization performance.

### B. Evaluation

The F1-score (fmeasure), precision, and recall on our test set are used as evaluations of results. As the data samples are unevenly distributed (the dataset is occupied by the "Other" category which fraction is 93%), in order to show the effectiveness of our method, the "Other Categories" label will be excluded when calculating the scores.

### C. Comparison model

We compare two versions of our model, with different numbers of sentence level Bi-directional LSTM layers or document level Bi-directional LSTM layers, against the baseline model, a two-layer non-hierarchical Bi-directional LSTM with attention mechanism (Bahdanau et al., 2014) [14]. All models have identical embedding layer, padding setting, dropout setting, training batch size and other hyper-parameters. All bidirectional LSTM layers have the same number of hidden units (168 *2).

## VI. RESULTS

### A. Test Scores

The scores on the test set are reported in Table 3. As can be seen from the Table, all hierarchical models significantly outperform the baseline (non-hierarchical).

The best model (2 sent-level and 3 doc-level layers) improved the F1-score (0.3560), precision (0.3382) and recall (0.4193) by 1377.2%, 3832.6% and 400.1% respectively.

## B. Sample of Extraction

A sample of extraction by our proposed hierarchical model from the test set is shown in Figure 2, from which we can see the model can extract distinguishable and relevant parts from standard lawsuit documents by correct labeling sentences in the documents.

We also build a demo of this work (http://demo.ikuquan.com/) which could be used to extract any civil document published on China Judgements Online.

Our model also has the ability to distinguish claims made from different parties, which provides more potentials for future tasks.

## C. Error Analysis

The result shows that extract the part of court's analysis remains a major challenge for our method. The model often focuses on one part (consists of several sentences), usually the main part, of the whole analysis. When we add attention mechanism to the final version of our model, it sometimes will ignore the entire analysis part.

We are intuitively believing the cause of this problem is that the semantics of the analysis part is more complex than the rest. In this part, the judge will use the syllogism approach to explain the reasons for the decisions of each claim. This makes the semantic differences between sentences unstable than other parts of the document. We will conduct more in-depth study of this issue in the future.

## VII. CONCLUSION AND FUTURE WORK

In this work, the result showed that Hierarchical Bidirectional LSTM models are effective tools for extracting information from real-life lawsuit document.

Our approach can serve as a basis for future work. First, because the criminal law-suit document and administrative lawsuit document have a different type of format, our modeling cannot be applied to these two kinds of document directly. But we can explore new models for them with similar algorithm and different training data. After our models cover all of the main legal areas, the information they extract will be helpful for judges and lawyers to catch on the case quickly.

Second, the extraction of information is a key step for the prediction of judge's decision on the similar case in the future. In America, as New York Times reported, some legal start-up companies have already had the product to predict the judge's decision, and we will develop a state-of-the-art system in Chinese version. Since there is increasing potential in the legal service market in China, we believe this technology will create huge commercial value and contribute to the academic community at the same time.

Claim (Party A)

一、判决被告阳光保险昆明公司、大地保险五华支公司在保险限额内赔偿原告周琴书损失176957元（残疾赔偿金44400元、误工费85438元、护理费28479元、营养费8500、后期医疗费3000元、伤残鉴定时产生的交通食宿费1255元、鉴定费3140元、精神抚慰金4000元），保险限额赔付不足部分，由被告徐其相、徐小意共同赔偿

First, the defendant, Sunshine Insurance Company compensate the plaintiff, in the limit within the insurance, Zhou QinShu's loss of 176,957 yuan (44,400 yuan disability compensation, loss of work costs 85438 yuan, care costs 28479 yuan, nutrition costs 8500, late medical expenses 3,000 yuan, disability identification when the traffic expenses 1255 yuan, identification fee of 3140 yuan, spiritual damages 4000 yuan), the exceed par paid by other defendant, Xu Qi and Xu Xiaoyi.

二、案件受理费由四被告承担

Second, the court costs borne by the four defendants.

Analysis

因然本案事故车辆系由被告徐小意所有，但本案中没有证据显示徐小意对本案事故的发生具有过错，故被告徐小意无需承担本案事故的赔偿责任

Although the owner of the accident vehicle is the defendant Xu Xiaoyi, but there is no evidence in this case shows Xu Xiaoyi was at fault for the accident, so the defendant Xu Xiaoyi no need to bear the case of the liability.

Decision

一、被告中国大地财产保险股份有限公司昆明市五华支公司于本判决生效后十五日内赔偿原告周琴书43128.57元

First, the defendant China Land Property Insurance Co., Ltd. Kunming Wuhua branch compensates plaintiff Zhou Qinshu 60837.84 yuan within 15 days after the entry into force of this judgment.

二、被告阳光财产保险股份有限公司昆明中心支公司于本判决生效后十五日内赔偿原告周琴书60837.84元

Second, the defendant Sunshine Property Insurance Co., Ltd. Kunming Center branch compensates plaintiff Zhou Qinshu 60837.84 yuan within 15 days after the entry into force of this judgment.

三、驳回原告周琴书的其他诉讼请求

Third, the plaintiff Zhou Qinshu's other claims are dismissed.

Fig. 2. A sample of extraction (case number: (2017) 黔0222锟斤拷锟斤拷3131锟斤拷) produced by our hierarchical bidirectional LSTM model on the test set.